CG-SSD: Corner Guided Single Stage 3D Object Detection from LiDAR Point Cloud

Ruiqi Ma[a,b], Chi Chen[a,b,]*, Bisheng Yang[a,b], Deren Li[a], Haiping Wang[a,b], Yangzi Cong[a,b], Zongtian Hu[a,b]

[a] State Key Laboratory of Information Engineering in Surveying, Mapping and Remote Sensing, Wuhan University, Wuhan 430079, China
[b] Engineering Research Center of Space-Time Data Capturing and Smart Application, the Ministry of Education of P.R.C., China

## Abstract

Detecting accurate 3D bounding boxes of the object from point clouds is a major task in autonomous driving perception. At present, the anchor-based or anchor-free models that use LiDAR point clouds for 3D object detection use the center assigner strategy to infer the 3D bounding boxes. However, in a real-world scene, the LiDAR can only acquire a limited object surface point clouds, but the center point of the object does not exist. Obtaining the object by aggregating the incomplete surface point clouds will bring a loss of accuracy in direction and dimension estimation. To address this problem, we propose a corner-guided anchor-free single-stage 3D object detection model (CG-SSD[1]). Firstly, the point clouds within a single frame are assigned into regular 3D grids. 3D sparse convolution backbone network composed of residual layers and sub-manifold sparse convolutional layers are used to construct bird's eye view (BEV) features for further deeper feature mining by a lite U-shaped network; Secondly, a novel corner-guided auxiliary module (CGAM) with adaptive corner classification algorithm is proposed to incorporate corner supervision signals into the neural network. CGAM is explicitly designed and trained to detect partially visible and invisible corners to obtains a more accurate object feature representation, especially for small or partial occluded objects; Finally, the deep features from both the backbone networks and CGAM module are concatenated and fed into the head module to predict the classification and 3D bounding boxes of the objects in the scene. The experiments demonstrate CG-SSD achieves the state-of-art performance on the ONCE benchmark for supervised 3D object detection using single frame point cloud data, with 62.77%mAP. Additionally, the experiments on ONCE and Waymo Open Dataset show that CGAM can be extended to most anchor-based models which use the BEV feature to detect objects, as a plug-in and bring +1.17%~+14.27%AP improvement.

## 1. Introduction

In recent years, autonomous driving technology, as one of the highlights of new energy vehicles, has been made great progress. 3D object detection is actively researched due to its importance for the perception module of autonomous driving. Generally, self-driving cars use sensors such as cameras, LiDAR, and radar to obtain data in the scene in real-time. The LiDAR sensor can work around the clock and is less sensitive to weather conditions (e.g., light, rain, and snow). Real-time object detection using LiDAR point clouds is one of the main research topics in 3D object detection tasks (Huang and Chen, 2020).

Learning effective point cloud features through neural networks is the premise of the 3D object detection task. However, different from continuous and dense 2D images, 3D point cloud data is sparse and unordered. Therefore, the excellent feature learning backbone (ResNet(He et al., 2016), VGG(Simonyan and Zisserman, 2015), ViT(Dosovitskiy et al., 2020)) in 2D image object detection task cannot be directly used in 3D point clouds. Then a new backbone for 3D point cloud features learning needs to be designed. MV3D(Chen et al., 2017) and AVOD(Ku et al., 2018) project the point clouds onto an image and then use a 2D convolutional network to learn the projected point clouds features. However, using projected images to express a full 3D point cloud will lose a lot of details. Feature learning methods represented by PointNet(Qi et al., 2017a) and PointNet++(Qi et al., 2017b) can directly consume unordered 3D point clouds and can extract useful features for point cloud semantics segmentation (Hu et al., 2019) and object detection (Lang et al., 2019; Shi et al., 2020a, 2019) tasks. Such point-based feature learning methods directly extract features from unordered point clouds without considering the spatial relationship between 3D points. Therefore, it will cause the problem of loss of point cloud spatial information. SECOND (Yan et al., 2018) proposed a method to divide the point clouds into regular-sized voxels, and then adopt 3D sparse convolution on the non-empty voxels to learn the point cloud features. While considering the spatial relationship between points, this method uses the sparsity between voxels to accelerate 3D convolution, which can achieve real-time detection. Since the voxel-based method has the advantages of fast calculation speed and strong point cloud features learning ability, many latest 3D object detection models use this method as the 3D backbone for point cloud features learning (Deng et al., 2021; Hu et al., 2021; Shi et al., 2020; Ye et al., 2020; Yin et al., 2021).

To better adapt the target assignment strategy in the 2D object detection task, most recent models choose to project the features to a bird's eye view (BEV) image after the 3D point cloud features learning. Similar to image object detection, existing models can be divided into anchor-based and anchor-free types according to the strategy selected by target assignment during model training. The anchor-based models(Chen et al., 2017; Deng et al., 2021; He et al., 2020; Lang et al., 2019; Shi et al., 2021a, 2020; Yan et al., 2018; Yang et al., 2018; Ye et al., 2020; You et al., 2019; Zheng et al., 2020; Zhou and Tuzel, 2018) obtain the object detection results by classifying and regressing pre-designed anchors of different sizes and orientations. Anchor-based models have good target size regression ability, but their localization capability of small targets is weak. The anchor-free models (Ge et al., 2021; Hu et al., 2021; H. Li et al., 2021; J. Li et al., 2021; Tang et al., 2021; Wang et al., 2020; Yang et al., 2020; Yin et al., 2021) obtain the 3D bounding box by detecting the key point and directly regressing the size and orientation of the objects. Anchor-free methods are good at small object detection. However, the regression ability of the size and direction of large targets is poor, and the results of vehicle object detection cannot compete with anchor-based methods.

Whether in 2D image or 3D point cloud object detection tasks, both anchor-based and anchor-free methods mentioned above all classify and regress the object based on the center point. The 2D image is composed of continuous and dense pixels, and the center point of the target is real, and the pixels around the center point are belonging to the target. The image object detection network can easily regress the position of the center point and the size of the object. However, the 3D point cloud data has sparsity and irregularity, and the acquired data is related to the real scene and the performance of the LiDAR sensor. The sensor can only obtain the reflection data of part target surface, and there is no point around the center of the 3D target. Especially for large targets such as vehicles, this phenomenon is more obvious.

To reduce the error caused by directly regressing the center point from the incomplete point cloud, CenterNet3D(Wang et al., 2020) model uses a detection head module to predict the four corners of the object in the BEV map and aggregates the predicted center point information to get the final result. AFDetV2(Hu et al., 2021) does the task of corner classification and only serves as an auxiliary module. And the corner classification results do not participate in the prediction of the boxes. The results of these two models illustrate that corner detection is beneficial for the 3D object detection task in point clouds. However, none of them consider the precise position of corners in the model, which can provide accurate boundary information for object localization and dimension regression. In addition, if the location of the corners can be obtained before the detection head module, the network can obtain more accurate 3D bounding boxes under the guidance of the position of the corners.

To improve the accuracy of regressing the center position and size of objects from the incomplete point clouds, we propose a corner-guided anchor-free single-stage 3D object detection model (CG-SSD). In the CG-SSD model, to allow the network to perceive more accurate target shape and orientation information, we innovatively added a corner-guided auxiliary module (CGAM) between BEV feature learning module and detection head module to perform supervised learning on corners that are difficultly observed by the LiDAR. The CG-SSD model achieved state-of-art performance with 62.77%mAP on the ONCE (Mao et al., 2021) testing set using a single frame point cloud. In addition, the results on the Waymo Open Dataset (Sun et al., 2020) and ONCE demonstrate the CGAM can be integrated with BEV map-based 3D object detection model as a plug-in, and present outstanding performance compared with the state-of-the-art methods. Our key contributions are as follows:

    1. A corner-guided anchor-free single-stage 3D object detection model is proposed, and corner detection is added to the inference of the network. The corner

detection module enhances the network's ability to detect targets with few LiDAR points, or small targets (*e.g.*, pedestrian, cyclist). The model achieves state-of-art results on the ONCE benchmark for supervised 3D object detection using single frame point cloud data.

2. A general corner detection auxiliary module is proposed, which can be used as a plug-in to enhance 3D detection models using a BEV map. The performance of the enhanced model is greatly improved, with +1.17%~+14.27%AP.

## 2. Related Work

According to the selection strategy of target assignment during network training, existing 3D object detection models can be divided into anchor-based and anchor-free models. The related research of these two models is described below.

## 2.1. Anchor-based 3D Object Detection Model

In the 2D image object detection task, to efficiently detect objects of different scales, previous works use pre-designed different anchors when training models (Girshick, 2015; Liu et al., 2015). Then networks classify the anchor and regress the difference between the detected target and the anchor. However, when using 3D point clouds for object detection tasks, it is difficult to detect objects directly from 3D point clouds due to the sparsity and discreteness of spatial point clouds. To solve this problem, Chen et al.(2017) propose a multi-view-based detection method, which projects the point clouds to the BEV and front-view to form three types of feature maps of height, density, and intensity, respectively. After the projection operation, the sparse and discrete 3D point clouds object detection problem is converted into a continuous and dense 2D image object detection problem, and then the anchor strategy is used to finally obtain the target in the 3D point clouds. PIXOR(Yang et al., 2018) adopts the same strategy to directly project the point clouds to the BEV, and then use the mature 2D convolutional network to learn the point cloud features and obtain the detection results. But the process of projecting the 3D point clouds directly onto an image loses a lot of information. And the different grid sizes of the projected image will directly affect the detection accuracy of small objects (pedestrians and cyclists).

VoxelNet(Zhou and Tuzel, 2018) proposes a single-stage 3D object detection method based on voxel feature extraction. The method first divides the point clouds into 3D grids, then uses PointNet(Qi et al., 2017a) to extract point features within voxels, and projects the features to BEV. Finally, a method similar to SSD(Liu et al., 2015) is used for 3D object detection. PointPillars (Lang et al., 2019) proposes a single-stage model for real-time 3D object detection. Different from VoxelNet(Zhou and Tuzel, 2018), this model divides the point clouds into pillars of the same size, and also uses PointNet(Qi et al., 2017a) to extract features. In the 3D feature extraction of these two models, the 3D sparse point clouds are divided into pillars or voxels, and then the PointNet(Qi et al., 2017a) feature extraction backbone that does not consider the spatial relationship is used, which will cause the loss of spatial information on $z-$ axis between points. When the 3D sparse point clouds are divided into voxels for feature extraction, since there are a large number of empty voxels in the detection space, using traditional 3D convolution will bring more computation. SECOND (Yan et al., 2018) proposes a 3D sparse convolution method for point cloud features learning. This model is similar to the VoxelNet(Zhou and Tuzel, 2018) network, but due to the addition of sparse convolution, it has the ability to detect 3D objects in real time. And the accuracy of the model has been further improved. The above methods all use the operation of projecting the point features to the BEV perspective after the 3D point cloud features extraction, which will bring the loss of spatial features.

Shi et al.(2020b) proposed a two-stage object detection method Part-$A^2$, which first uses an anchor-based method to obtain objects' proposal on first stage. Then use the 3D RoI Pooling method to extract the points in each proposal, and use 3D spare convolution to extract features to do the 3D box refinement. Shi et al.(2021, 2020a) propose PV-RCNN and PV-RCNN++ to improve the Part-$A^2$ method. When the model extracts 3D features, both voxel-based and point-based features are extracted to optimize the bounding box in the second stage. Although these two-stage models have achieved some advantages in accuracy, their models are more time-consuming as the amount of computation increases.

To get more accurate detection results, some other models achieve better results by adding additional supervision and assistance to the detection head. For example, SA-SSD(He et al., 2020) adds an adopted PSRoIAlign(Dai et al., 2016) to the model classification stage, which makes the model obtain higher 3D accuracy. Zheng et al. (2020) proposed the CIA-SSD model, which added an Intersection over Union(IoU) prediction branch to filter out error boxes with high classification scores and low IoU values and improve the accuracy in difficult categories in the KITTI dataset(Geiger et al., 2013). The recent Voxel-RCNN(Deng et al., 2021) model is an efficient anchor-based two-stage method. This model is a variant of the PV-RCNN model. Unlike the PV-RCNN model, which simultaneously extracts voxel and point features, it only extracts voxel features during 3D feature extraction, which improves the performance of the model. Miao et al.(2021) proposed an anchor-based model that combines point, voxel, and BEV feature expression to extract favorable features to improve object detection results.

## 2.2. Anchor-free 3D Object Detection Model

Since anchor-based models need pre-designed anchors and use models to predict the difference between the ground truth box and the anchor, the target output values of the model are all within a certain range, which is more conducive to network convergence. However, in the multi-object detection task, there is a large difference between the sizes of the targets, and the manual design of anchors shows limitations. To adapt multi-scale objects, anchor-free models are proposed in the task of 2D image object detection(Y. Chen et al., 2020; Duan et al., 2019; Huang et al., 2015; Law and Deng, 2020; Yang et al., 2019). In 3D object detection, Shi et al. (2019) propose an anchor-free two-stage model PointRCNN. The model uses the PointNet++(Qi et al., 2017b) to extract point features and perform semantic segmentation, and simultaneously regress the corresponding 3D bounding box for each foreground point. Then RoI pooling is used to extract the points in each proposal, and the proposal is optimized in the second stage to obtain the final detection result. Compared with the previous anchor-based method, this method greatly improves the detection accuracy of the cyclist category. However, since PointNet++(Qi et al., 2017b) is used to extract features in PointRCNN(Shi et al., 2019) when it performs multi-scale feature learning, some small object features may be lost during the sampling process, resulting in low accuracy in pedestrian detection. Similar to the anchor-free model in the 2D object detection task(Duan et al., 2019), Q. Chen et al. (2020) proposed a simple and efficient single-stage anchor-free model, which is similar in structure to the SECOND(Yan et al. et al., 2018). The model regards the object as an independent Hotspot on the BEV map and obtains the 3D bounding box by classifying each Hotspot and regressing its corresponding size and orientation angle. Benefiting from the advantages of the anchor-free model in small target detection, the accuracy of this method in pedestrian and cyclist detection is the best at that time. The subsequent optimized models (Ge et al., 2020, 2021; J. Li et al., 2021; Tang et al., 2020; Wang et al., 2020) all show comparable performance to anchor-based or two-stage methods even better detection results.

Compared with the anchor-based model, the anchor-free model shows greater advantages in 3D object detection, especially in the detection of small objects (pedestrian, cyclist). Whether it is an anchor-based model or an anchor-free model, they all design the network through the strategy of center point assignment. However, in the special data type of sparse point clouds, the center point of the object does not exist, so the network learning "virtual point" will inevitably lead to inaccurate prediction. Therefore, we designed a corner-guided single-stage detection model. By adding a corner auxiliary supervision module, the network can perceive more accurate target edge and orientation information to obtain higher detection accuracy.

## 3. Method

In this section, we introduce the CG-SSD model proposed in this paper. CG-SSD consists of voxelization, 3D and 2D feature learning, CGAM, and detection head. First, we assign the point clouds into the regular 3D grids and extract voxel features using 3D sparse convolution. After getting the 3D voxel features, we project the features to the BEV and use the 2D backbone to extract deeper features. Unlike other networks, we add CGAM between the 2D backbone and detection head module. Then, the corner features extracted by the auxiliary network and the features obtained by the 2D backbone are fused, and finally, the classification and regression tasks of the object are performed. Figure 1 shows the architecture of the CG-SSD. Next, we will introduce each part of the CG-SSD separately.

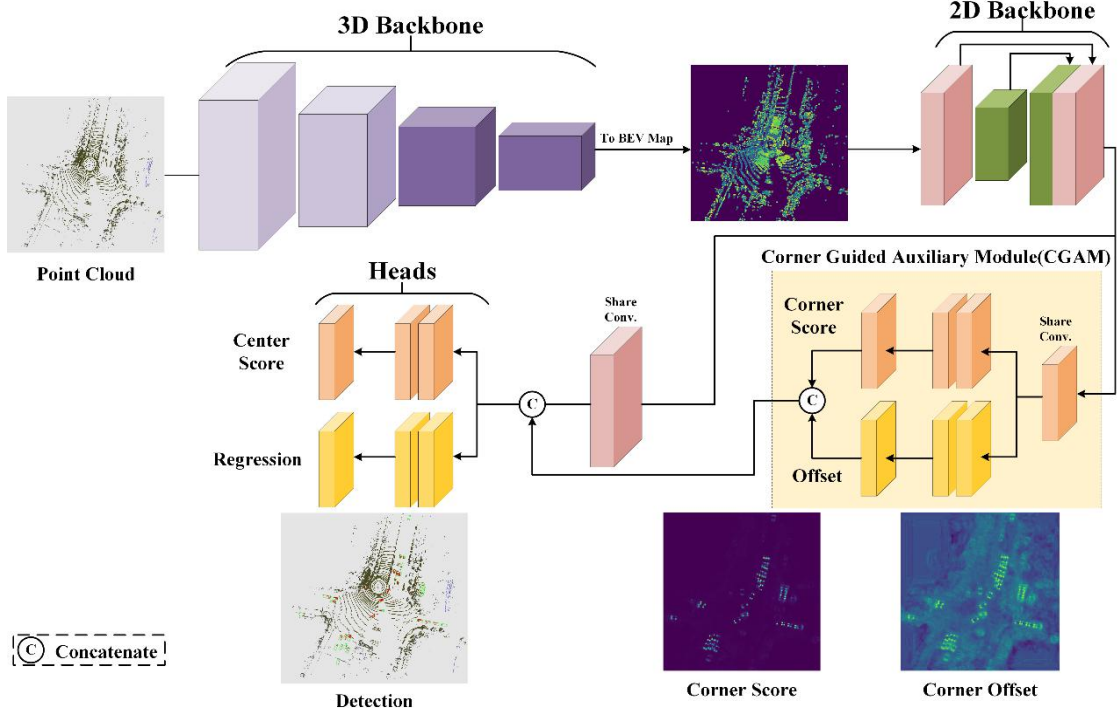

Figure 1. The structure of the CG-SSD Module

## 3.1. Backbone

**Voxelization.** To deal with sparse and irregular point clouds data, there are mainly methods based on point, voxel, and projection to learn point cloud features. In this paper, the voxel-based feature learning method is selected that is efficient and considers the spatial neighborhood relationship. Given a point cloud dataset $P = \{p_0, p_1, \cdots, p_n\} \subseteq \mathcal{R}^{3+m}$, point cloud range $[x_{min}, y_{min}, z_{min}, x_{max}, y_{max}, z_{max}]$ and voxel size $[v_x, v_y, v_z]$, the voxel index of a point $p_i = \{x_i, y_i, z_i, c_0, \cdots c_m\}$ within point cloud range is located according to Equation 1, where $c$ is the attribute of point and $m$ is the dimension of point cloud feature.

$$Index_i = \left[int\left(\frac{x_i - x_{min}}{v_x}\right), int\left(\frac{y_i - y_{min}}{v_y}\right), int\left(\frac{z_i - z_{min}}{v_z}\right)\right] \subseteq \mathcal{R}^3 . \tag{1}$$

Due to the sparsity of the point clouds, the number of points within each voxel is different. Thus we set a fixed value *T*. If the number of points belonging to the voxel is greater than *T*, then *T* points are randomly sampled from all the points of the voxel. If the number of points is less than *T*, we pad to *T* with zeros. After such an operation, we get $N$ no-empty voxels $V$, forming a three-dimensional tensor: $N \times T \times (3 + m)$. We will use 3D sparse convolution to perform feature extraction on this Tensor.

**3D backbone.** To efficiently learn the features of 3D voxels. We adopt 3D Sparse Convolution(Yan et al., 2018) as the 3D backbone of CG-SSD. The structure of this module is shown in Figure 2, which is a backbone composed of residual layers and sub-manifold sparse convolutional layers. After the multi-layer convolutions, each voxel can learn spatial features at different scales, and finally, obtain voxels with richer features. In this paper, after three convolutions with stride = (2, 2, 2), kernel = (3, 3, 3), padding = (1, 1, 1), the size of each voxel becomes $[8 * v_x, 8 * v_y, 8 * v_z]$. To simplify object classification and regression tasks, 3D voxel features are projected to BEV according to the spatial indices of voxels. And in order to reduce the amount of calculation, the 3D sparse convolution with stride = (2,1,1), kernel = (3, 1, 1), padding = (0, 0, 0) is finally used to reduce the resolution in the $z-$axis. The final BEV feature map is $I \subseteq \mathcal{R}^{H \times W \times (D \times F)}$, where $H, W$ is the size of the projected feature map. Since the feature map is obtained by 3D voxel projection, in order to avoid feature loss, the features in the $z-$axis are connected, and the final feature dimension is $D \times F$.

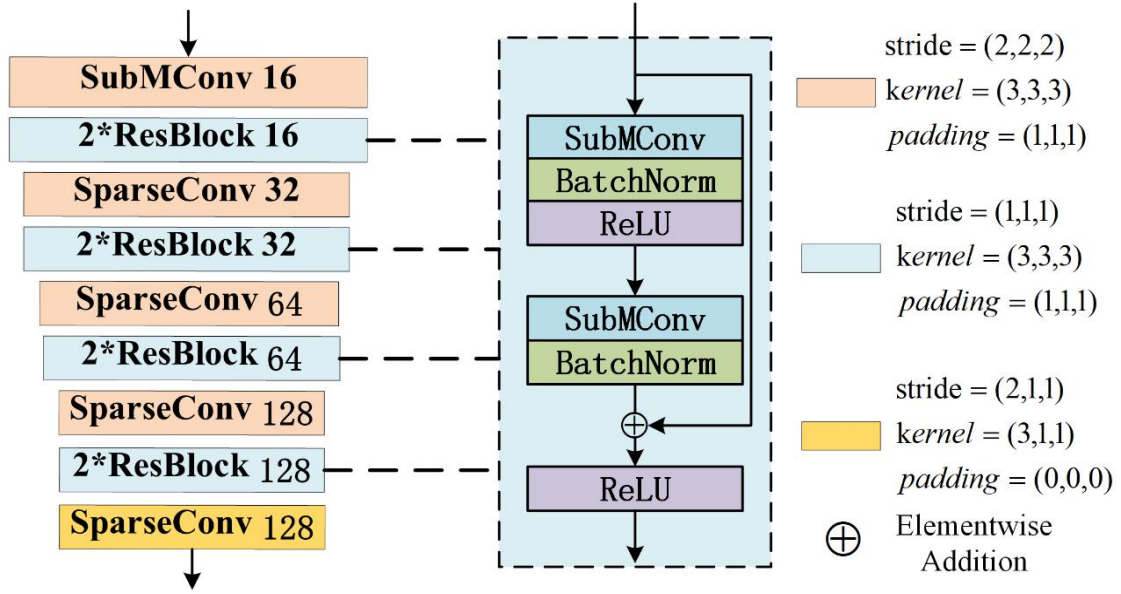

Figure 2. 3D Sparse Residual Convolution Backbone

**2D backbone.** After obtaining the BEV feature map, we also adopted the same learning strategy as other state-of-art models(Hu et al., 2021; Shi et al., 2020; Yin et al., 2021), that is, using the 2D convolution layers to learn deeper features. Here we design an SSD-like(Liu et al., 2015) network as the 2D backbone. As shown in Figure 3, the 2D backbone consists of multiple sets of Conv2D-BatchNorm2D-ReLU (CBR) and ConvTranspose2D-BatchNorm2D-ReLU (CTBR) block. First, the CBR blocks with stride = 1, kernel = 3, and channel = 128 are adopted, focusing on learning the characteristics of small objects. Second, by using the CBR block with stride = 2, kernel = 3, and channel = 256, the resolution of the feature map is reduced and the receptive field of the convolution kernel is increased, to learn the features of the large object. Third, we use 2D deconvolution with stride = 1, 2 and channel = 256, 256 respectively, so that the size of the output 2D features is the same, and the final output of the 2D backbone is the combination of the output deconvolution layers. Through the 2D backbone, the network can learn different levels of features, which can be better used for object classification and detection of different sizes.

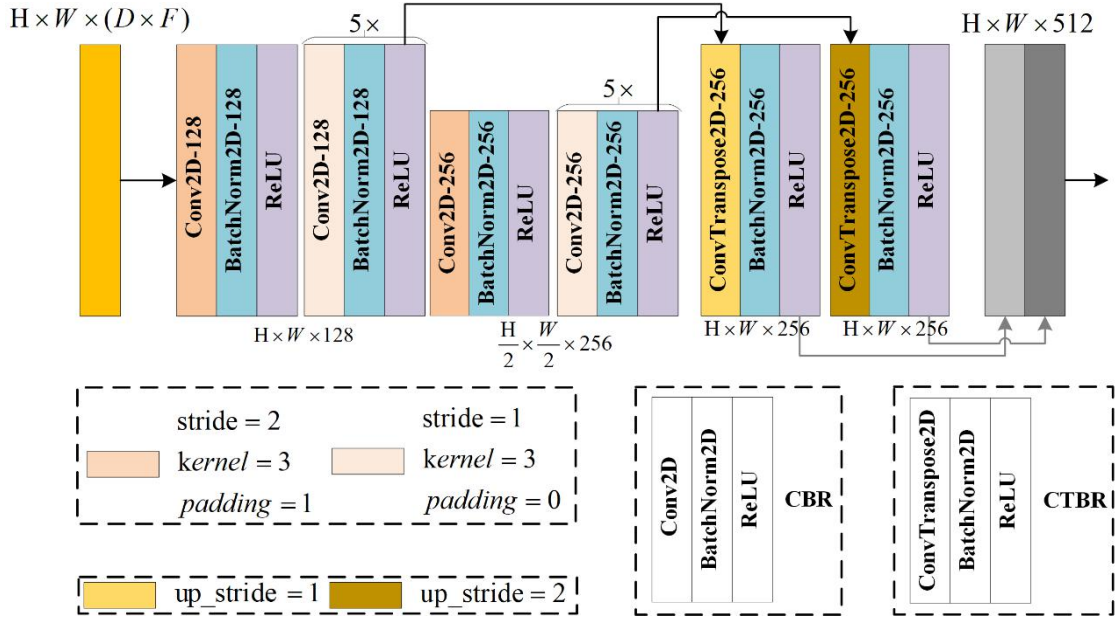

Figure 3. The 2D backbone of CG-SSD

## 3.2. Corner Guided Auxiliary Module (CGAM)

Most of the recent models directly perform object classification and regression after 3D and 2D feature learning. However, due to the influence of factors such as occlusion between objects and the distance from the target to the sensor, it is difficult for the sensor to obtain complete target surface point clouds. And it is unfavorable for the network to learn an object center point from incomplete point clouds. Therefore, CG-SSD proposes to supervise the network to learn more accurate center points by predicting the corners of the 3D bounding boxes on the BEV map, thereby improving the detection accuracy of the network.

### 3.2.1. Corner Selection Strategy

As shown in Figure 4, in the BEV map, if the sensor can observe the object, it must be able to observe a corner (visible corner) of its bounding box. Since many object point clouds are distributed around the corner, the network can easily learn the position of the corner. However, for the other three corners (partly visible corner l, partly visible corner w, and invisible corner), which are hard to be observed by the sensor, they should be given more attention. By learning the positions of these corners, the model is potentially guided to aggregate the target features towards the center point.

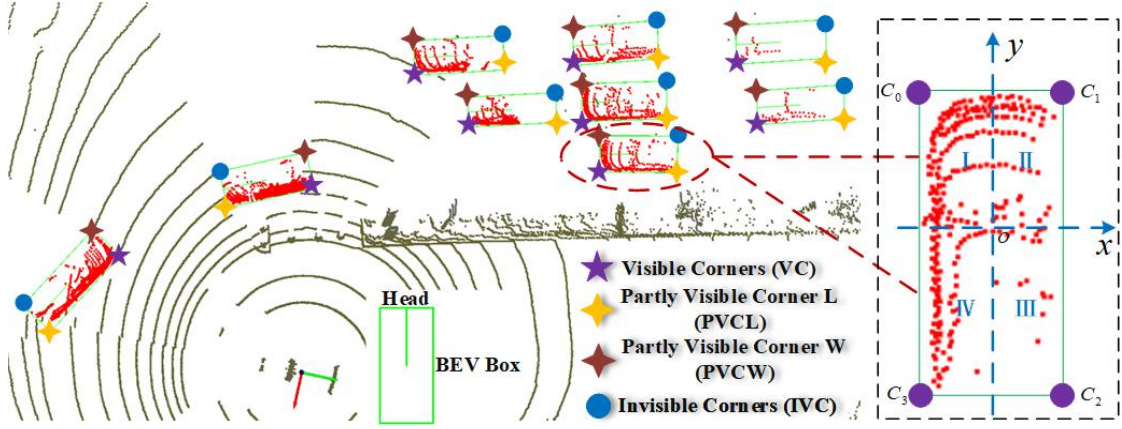

Figure 4. The visibility of object corners in the BEV map

Algorithm 1. The adaptive corner selection algorithm

**Data:** $n$ points with local coordinate $P \subseteq R^{n*3}$, points in 3D bounding boxes index $P_{index} \subseteq R^{n*1}$, 3D bounding boxes number $m$, corners of 3D bounding boxes' top surface $C \subseteq R^{m*4*3}$, index used to select auxiliary corners
$$index = [(2,3,1),(3,2,0),(0,1,3),(1,0,2)]$$
**Result:** selected corners $C_f, C_l, C_w$ for auxiliary module

1    $C_f \leftarrow zeros(m, 2)$;
2    $C_l \leftarrow zeros(m, 2)$;
3    $C_w \leftarrow zeros(m, 2)$;
4    **for** $i \leftarrow 0$ **to** $m$ **do**
5       $p_i \leftarrow p_{index} == i$;
6       $X \leftarrow P[p_i, 0]$;
7       $Y \leftarrow P[p_i, 1]$;
8       $q0 \leftarrow sum(X < 0 \text{ and } Y > 0)$;
9       $q1 \leftarrow sum(X > 0 \text{ and } Y > 0)$;
10      $q2 \leftarrow sum(X > 0 \text{ and } Y < 0)$;
11      $q3 \leftarrow sum(X < 0 \text{ and } Y < 0)$;
12      $q \leftarrow [q0, q1, q2, q3]$;
13      $sub\_q \leftarrow [(q0+q1+q3),(q1+q0+q2),(q2+q1+q3),(q3+q2+q0)]$;
14      $valid_q \leftarrow sum(q > 0)$;
15      **if** $valid_q <= 2$ **then**
16         $max_i \leftarrow argmax(q)$;
17      **else**
18         $max_i \leftarrow argmax(sub\_q)$;
19      **end**
20      $C_f[i] \leftarrow C[i, index[max_i, 0]]$;
21      $C_l[i] \leftarrow C[i, index[max_i, 1]]$;
22      $C_w[i] \leftarrow C[i, index[max_i, 2]]$;
23 **end**

The ground truth 3D bounding boxes of the datasets (KITTI, ONCE, Waymo) are composed of the center point position, dimension (length, width, height), and rotation angle. To supervise the network to learn corner features, during the training process, we need to calculate the position of each visible corner and the other three corners. However, because the orientation of the target in the LiDAR coordinate system is different, and the density of points in the target box is different, the location of each category corner cannot be determined. Therefore, we propose an adaptive corner selection algorithm based on the number of points in the 3D bounding box. First, we convert the LiDAR coordinate system to the local coordinate system through Equation 2, to obtain the points' coordinate in the 3D bounding box in local coordinate.

$$\begin{vmatrix} x_i' \\ y_i' \\ z_i' \end{vmatrix} = \begin{vmatrix} x_i - x_c \\ y_i - y_c \\ z_i - z_c \end{vmatrix} \begin{vmatrix} \cos(\theta) & -\sin(\theta) & 0 \\ \sin(\theta) & \cos(\theta) & 0 \\ 0 & 0 & 1 \end{vmatrix} \quad i = 0,1,2,\cdots,m \qquad (2)$$

where $(x_i, y_i, z_i)$ is coordinate in LiDAR coordinate system, $(x_c, y_c, z_c)$ is object center in LiDAR coordinate system, $\theta$ is object orientation angle, $m$ is the number of points, and $(x_i', y_i', z_i')$ is points' coordinate in local coordinate. As shown on the right of Figure 4, the center of the object is taken as the origin, the front direction of the object is the positive direction of the $y-$ axis, and the positive direction of the $x-$ axis is the direction in which the $y-$ axis rotates 90° clockwise. As described in Algorithm 1, according to the $x-y$ axis, we divide the object's points into four quadrants and count the number of points in each quadrant. The corner corresponding to the quadrant with the most points is visible (VC in Figure 4), and it is the easiest to detect by the network, while the remaining three corners are likely to be observed by the sensor. The corner opposite to the visible corner is the invisible (IVC in Figure 4), and the other two corners are partly visible (PVCL and PVCW in Figure 4).

### 3.2.2. Auxiliary Module

In the previous model(Hu et al., 2021; Yin et al., 2021), after obtaining more advanced features through the 2D backbone, the head module is used for object detection. To obtain more accurate detection results, we added a corner-guided auxiliary module (CGAM) before the head module (Figure 5). The network learns the edge information and rotation information of the object by detecting partially visible and invisible corners and obtains a more accurate center point position, dimension, and angle. This module is mainly composed of a CBR block, which is supervised by calculating the losses of classification and position offset during network training.

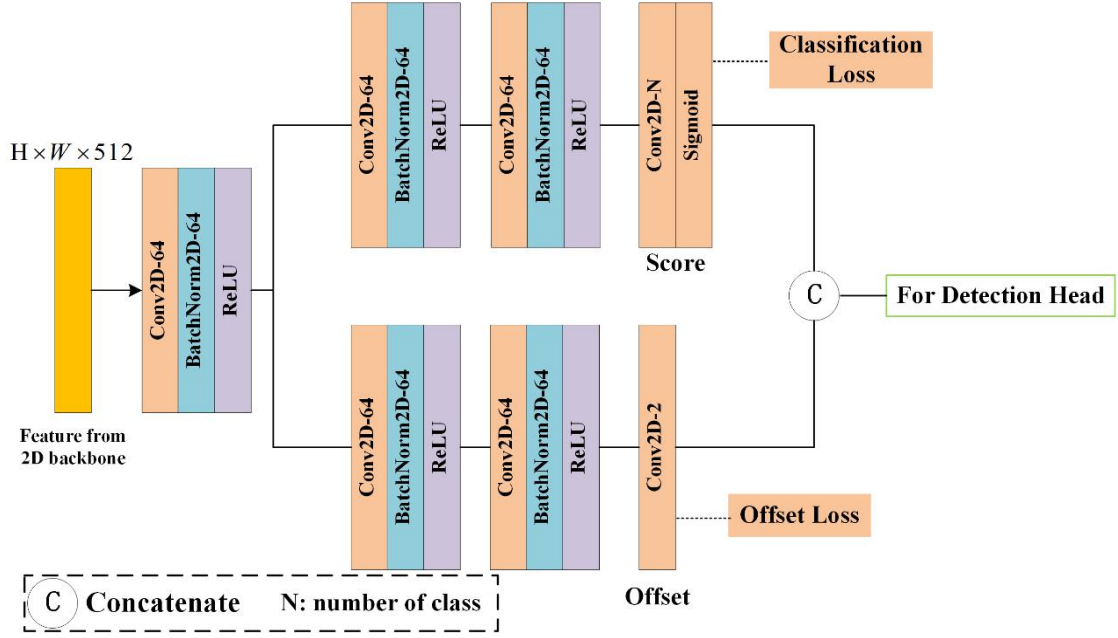

Figure 5. Corner Guided Auxiliary Module

In the classification task of the CGAM, the size of the final output of the model is $H \times W \times C$, where $H$ and $W$ are the sizes of the score map, and $C$ is the number of classes. For each object, only the pixel where the corner is located is positive, and the rest are negative. The target assignment strategy will bring about an unbalanced distribution of positive and negative samples, which affects network training. Therefore, we adopt the same strategy as CornerNet(Law and Deng, 2020), that is to reduce the penalty of pixels near the corners. The 2D Gaussian value (Equation 3) is set for the pixel which is located within a circle with the corner as center and $r$ as radius.

$$p = e^{-\frac{x^2+y^2}{2\sigma^2}} \qquad (3)$$

where $x$ and $y$ are the coordinate differences between the pixel and corner in BEV map, $\sigma = \frac{r}{3}$, and $p$ is the target value. In this paper, $r$ is empirically set to 2 as the CornerNet(Law and Deng, 2020).

In the regression task of the auxiliary module, the final output of the model is $H \times W \times 2$, whose third dimension represents the offset between the pixel center's coordinate and the corner's coordinate on the BEV map. Due to the voxel size setting, row and column numbers of the corner in the BEV map are decimals. Therefore, there is an offset when using the row and column numbers to express the precise position of the corner. And the offset can be calculated according to Equation 4-6. The regression branch in CGAM is used to predict the deviation value.

$$p_x, p_y = \left( \frac{x_c - x_{min}}{v_x \times out_{factor}}, \frac{y_c - y_{min}}{v_y \times out_{factor}} \right) \qquad (4)$$

$$x_o = x_c - [int(p_x) \times v_x \times out_{factor} + x_{min}] \qquad (5)$$

$$y_o = y_c - [int(p_y) \times v_y \times out_{factor} + y_{min}] \qquad (6)$$

where $x_c$, $y_c$ are the corner's value in LiDAR coordinate system, $x_{min}$, $y_{min}$ are the minimal value of inputted point clouds, $v_x$, $v_y$ are the voxel size, $out_{factor}$ is the ratio of the voxel size before and after the 3D backbone, and is 8 in this paper. The $x_o$ and $y_o$ are the target value.

Additionally, the score map, regression map, and feature from the 2D backbone are concatenated for the detection head module. The combined feature has a channel with $f + C \times n + 2 \times n$, where $C$ is the number of predicted classes, $n$ is the number of corners for each object, $f$ is the number of features obtained by the 2D backbone and set to 512, and 2 represents the corner offset value.

## 3.3. Detection Head and Loss Function

### 3.3.1. Detection Head

The detection head module is designed for classification and regression tasks with enhanced features. In the classification branch, the strategy of center assignment and Equation 3 are used to obtain the ground-truth value of network training. In the regression branch, the model outputs the offset of the center point, the dimension, and the orientation angle of objects, respectively. And only the pixel corresponding to the center point position is assigned the true value, and other pixels are assigned zeros. In the process of network training, we get such true value according to Equation 7.

$$\begin{aligned} x_t &= x_o, y_t = y_o, z_t = z, \\ w_t &= \log(w), l_t = \log(l), h_t = \log(h), \\ a1 &= \sin(\theta), a2 = \cos(\theta) \end{aligned} \qquad (7)$$

where $x_o$ and $y_o$ are the offset value of the object's center, z is the center point's z-axis value, $(w, l, h)$ are the dimension (width, length, height), and $\theta$ is orientation angle. Figure 6 shows the detection head module of the CG-SSD model, which is composed of multiple CBR layers. First, a shared CBR layer transforms the features from the 2D backbone. Second, both the classification and regression branches consist of two middle CBR layers and a Conv2D layer to get the predictions of the network. The input of the classification and regression branches is the concatenation of the features learned by the share layer and the head with the same predicted category in the auxiliary module. In this module, the stride of all CBR layers is 1, the kernel is 3, and the out channel is 64.

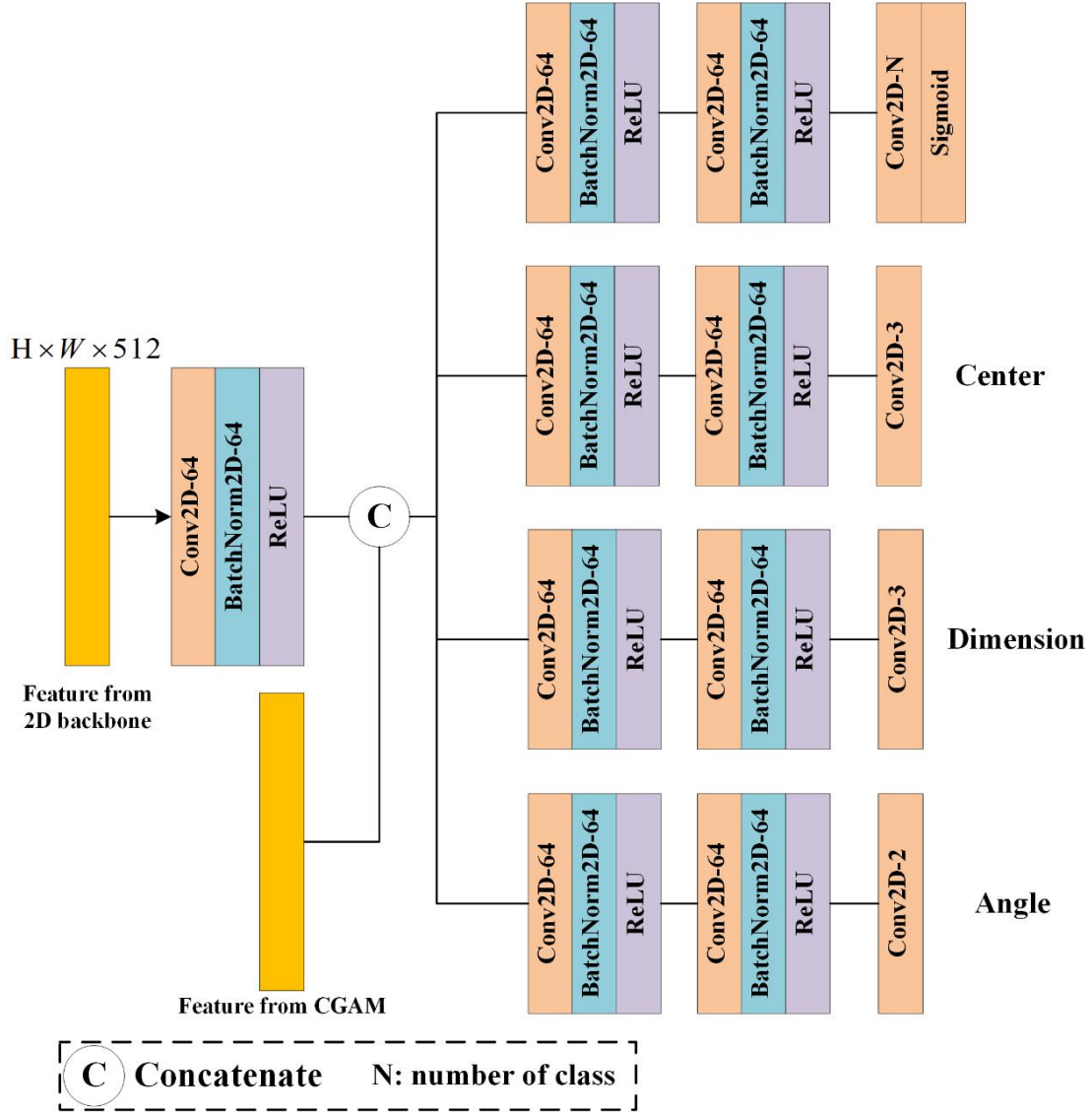

Figure 6. Head module of CG-SSD

### 3.3.2. Multi-task Loss

In the CG-SSD model, the classification loss function is a variant of Focal Loss(Law and Deng, 2020), and the regression loss function is $L_1$ Loss. The focal loss can be calculated by Equation 8.

$$L_f = \frac{-1}{N_c}\sum_{c=1}^{C}\sum_{i=1}^{H}\sum_{j=1}^{W}\begin{cases}(1-p_{cij})^\alpha \log(p_{cij}) & if\ y_{cij}=1 \\ (1-y_{cij})^\beta (p_{cij})^\alpha \log(1-p_{cij}) & otherwise\end{cases} \quad (8)$$

where $N_c$ is the number of positive samples, $C$ is the number of classes, $H$ and $W$ are the sizes of score map, $y_{ij}$ is the ground truth value of classification, the $p_{cij}$ is the network prediction, and $\alpha$, $\beta$ are the hyperparameters which are set to 2, 4 respectively.

The $L_1$ Loss is used to the regression module in the CG-SSD model, which can be calculated by Equation 9,

$$L_g = \frac{1}{N_r}\sum_{i=1}^{N_r}\left|G_i^t - G_i^p\right| \qquad (9)$$

where $N_r$ is the number of positive pixels in regression map, $G^p$ is the prediction of the regression branch, and the $G^t$ is the ground truth value. Especially, in the auxiliary module, $G^t \in [x_o, y_o]$ and $x_o, y_o$ are the offset value of corner from Equation 5-6. And in the detection head module, $G^t \in [x_t, y_t, z_t, w_t, l_t, h_t, a_1, a_2]$ which is calculated from Equation 7.

The total loss $L$ of CG-SSD model is:

$$L = L_{center\_cls} + \gamma \times L_{center\_reg} + \lambda \times L_{aux} \qquad (10)$$

$$L_{aux} = L_{corner\_cls} + L_{corner\_reg} \qquad (11)$$

where $L_{center\_cls}$ is focal loss of detection head module, $L_{center\_reg}$ is the L1 loss of regression branch in detection module, $L_{aux}$ is the loss of the corner guided auxiliary module, which is the sum of the classification and regression loss of the corner, and $\gamma$, $\lambda$ are the loss weight, which is set to 0.25, 0.25 respectively.

## 4. Experiments

### 4.1. Dataset

The CG-SSD model is tested on the ONCE Dataset (Table 1). ONCE dataset was published for 3D object detection in the autonomous driving scenario, consisting of 1 million LiDAR scenes. And the LiDAR data is acquired by a 40-beam LiDAR sensor in multiple cities in China. ONCE dataset contains 6 training sequences (5k scenes), 4 validation sequences (3k scenes), and 10 testing sequences (8k scenes) for model evaluation. During the evaluation, it merges the car, bus, and truck into a super-class "vehicle". The official evaluation metric for ONCE dataset is orientation-aware mean average precision (mAP), which is obtained by averaging the AP of 3 categories. In this paper, we set a detection point cloud range of $[-75.2, 75.2]$ for $x$ and $y$ axis, and $[-5.0, 3.0]$ for $z$ axis.

Table 1. The dataset used in CG-SSD model

| Dataset | Publish Year | Scenes | Classes | Area |
|---|---|---|---|---|
| ONCE | 2021 | 1M | car,bus,truck,pedestrian,cyclist | China |
| Waymo | 2019 | 230k | vehicle,pedestrian,cyclist,sign | US |

To demonstrate that CGAM can be integrated as a plug-in to most networks using BEV features, we also evaluated it on the Waymo Open Dataset (Table 1). Waymo contains 798 training sequences, 202 validation sequences, and 150 testing sequences with vehicles, pedestrians, and cyclists. The LiDAR data is captured with a 64-beam LiDAR sensor. In this paper, we only predict the object in the point cloud range of $[-75.2, 75.2]$ for $x$ and $y$ axis, and $[-2.0, 4.0]$ for $z$ axis. Following the open-source project OpenPCDet[2], we only train the model on 20% of the training set and evaluate it on the full validation set.

## 4.2. Network Details

**Data Augmentation.** Following SECOND(Yan et al., 2018), we apply scaling, rotation, and flipping on individual objects and database sampling for every frame data. Firstly, to keep a more balanced number of categories in the training frame, we randomly sample 1, 4, 3, 2, 2 objects' point clouds for car, bus, truck, pedestrian and cyclist respectively from the all-training ground truth 3D bounding boxes' point clouds and past them into the current frame. Then, the random flipping along $x$, $y$-axis, random rotation of $\left[-\frac{\pi}{4}, \frac{\pi}{4}\right]$ and scaling of $[0.95, 1.05]$ are respectively applied to the objects' boxes and frame's point clouds. In the experiments of verifying that our auxiliary module can be made a plug-in in other models, we follow the original model's data augmentation and do not change anything.

**Optimization.** We train the CG-SSD with batch size = 30 for 80 epochs on five Nvidia RTX 3090 GPUs in ONCE dataset. The AdamW(Loshchilov and Hutter, 2019) optimizer with a one-cycle policy is used during training our CG-SSD model. The max learning rate is set to 0.0028, the division factor is set to 10, the momentum range is set from 0.95 to 0.85, and the weight decay is set to 0.01.

**Evaluation.** For ONCE and Waymo datasets, the 3D IoU based metric average precision is used for evaluating different models. The IoU between predicted boxes $B_p$ and ground truth boxes $B_{gt}$ can be calculated using the following metric:

$$\text{IoU} = \frac{area(B_{gt} \cap B_p)}{area(B_{gt} \cup B_p)}. \tag{12}$$

In ONCE dataset, it uses the IoU threshold (0.7, 0.7, 0.7, 0.3, 0.5 for car, bus, truck, pedestrian, and cyclist) to determine whether the predicted box is positive. And the predicted orientation angle is added to filter the positive boxes if the angle can fall into the $\pm 90°$ range of the matched ground-truth box. The official metric calculates the AP@50 by 50 score thresholds with the recall rates from 0.02 to 1.00 at step 0.02. And it merges the car, bus, truck into a super-class named vehicle. The mean AP (mAP) is reported by the official metric which is obtained by averaging the AP of 3 classes (vehicle, pedestrian, and cyclist).

In Waymo dataset, it uses the same method as the ONCE to evaluate the model's performance. The official metric of Waymo dataset provides AP and APH scores. For APH metric, it is calculated by weighting the orientation angle accuracy to the AP. And the orientation angle accuracy is defined as $\min(|\tilde{\theta} - \theta|, 2\pi - |\tilde{\theta} - \theta|)/\pi$, where $\theta$ and $\tilde{\theta}$ are the angle of the ground-truth box and predicted box. The IoU

---

[2] https://github.com/open-mmlab/OpenPCDet

threshold of Waymo dataset is set to 0.7, 0.5, 0.5 for the vehicle, pedestrian, and cyclist views respectively. Additionally, the performance of two difficulty levels is also calculated by the official evaluation metric. "LEVEL 1" boxes have more than five LiDAR points, and "LEVEL 2" have at least one.

### 4.3. Results and Analysis

### 4.3.1. Evaluation Using the ONCE Testing Set

Table 2 shows the result of our CG-SSD module on the testing set of ONCE dataset. Our method achieves the state-of-the-art results on mAP with only using one frame LiDAR point cloud. On ONCE testing set, our model achieves 68.00%AP, 52.81%AP, and 67.50%AP for the vehicle, pedestrian, and cyclist categories, and 62.77%mAP for three categories. It outperforms all models in the benchmark by at least 1.5%mAP.

Compared with the anchor-free model(CenterPoints(Yin et al., 2021)), our CG-SSD gets +1.53%mAP better than it. Compared with the anchor-based models (PointPillars(Lang et al., 2019), SECOND(Yan et al., 2018), PV-RCNN(Shi et al., 2020)), CG-SSD has the better performance on the pedestrian and cyclist with at least +30.15%AP and +5.57%AP. The anchor-based models have better performance (+8.98%AP) in vehicle detection overall anchor-free models, while their performance drops significantly in pedestrian (-30.15%AP) and cyclist(-5.57%AP) detection.

The main reason that leads to this phenomenon, is the limited size of the BEV grids. The pedestrian and cyclist can only occupy one or two pixels in the BEV feature map. So, it is difficult for the network to predict its dimension and orientation angle. Benefitting from the advantages of the proposed auxiliary module for partly visible corners detection, our model shows a larger margin than anchor-based models, especially in the pedestrian and cyclist categories.

Table 2. Performance comparison on the ONCE testing set

| Model | Stage | Anchor-free | AP@50 Vehicle | Pedestrian | Cyclist | mAP (%) |
|---|---|---|---|---|---|---|
| PointPillars*(Lang et al., 2019) | One | No | 69.52 | 17.28 | 49.63 | 45.47 |
| SECOND*(Yan et al., 2018) | One | No | 69.71 | 26.09 | 59.92 | 51.90 |
| PointRCNN*(Shi et al., 2019) | Two | Yes | 52.00 | 8.73 | 34.02 | 31.58 |
| PV-RCNN*(Shi et al., 2020) | Two | No | **76.98** | 22.66 | 61.93 | 53.85 |
| CenterPoints*(Yin et al., 2021) | One | Yes | 66.35 | 51.80 | 65.57 | 61.24 |
| CG-SSD(ours) | One | Yes | 68.00 | **52.81** | **67.50** | **62.77** |

Note: *: represents the result of the model is from ONCE Benchmark.

### 4.3.2. Evaluation Using the ONCE Validation Set

On the ONCE benchmark server, we can only get the AP and mAP of the three categories. To analyze the CG-SSD's performance in the different distances of an object to the sensor, we also evaluate our CG-SSD model on the ONCE validation set. As shown in Table 3, the detection precision of our model outperforms CenterPoints(Yin et al., 2021) for all objects in different distance ranges. And we visualize the detection result of CenterPoints(Yin et al., 2021) and our CG-SSD model in Figure 7.

For vehicle detection, CG-SSD has a slight advantage (+0.12%AP) over CenterPoints(Yin et al., 2021) in $0-30m$ range, and has a significant preference (+1.68%AP & +1.38%AP) in $30-50m$ and $50m-inf$ range. The anchor-based models(PointPillars(Lang et al., 2019), SECOND(Yan et al., 2018), PV-RCNN(Shi et al., 2020)) show better performance than CG-SSD (+0.97%AP). The anchor-free model needs to directly predict the object from the incomplete object point clouds, so it is difficult to accurately estimate the dimension and orientation value of the target. However, the anchor-based models regress the difference between the designed anchor and the ground truth box. Even if the predicted value is not accurate enough, the anchor-based models can get results closer to ground truth boxes than the anchor-free models.

And for pedestrian and cyclist detection in $50m-inf$, CG-SSD can get at least +1.39%AP than CenterPoints(Yin et al., 2021). The results illustrate that as the distance of the object from the sensor increases, CG-SSD is more stable and more accurate in detecting vehicles than the CenterPoints(Yin et al., 2021) model. This is mainly because as the distance increases, the number of points of the target decreases, and the surface contour of the target becomes more blurred. While the center-based model gets the dimension and orientation of the object by aggregating edge features to the center point, it is difficult to learn the accurate 3D information of the object with limited points. However, in the CG-SSD model, it learns partly visible and invisible corners' point information, and then fuses corners' features into the object detection task as a way for the model to perceive more accurate 3D information.

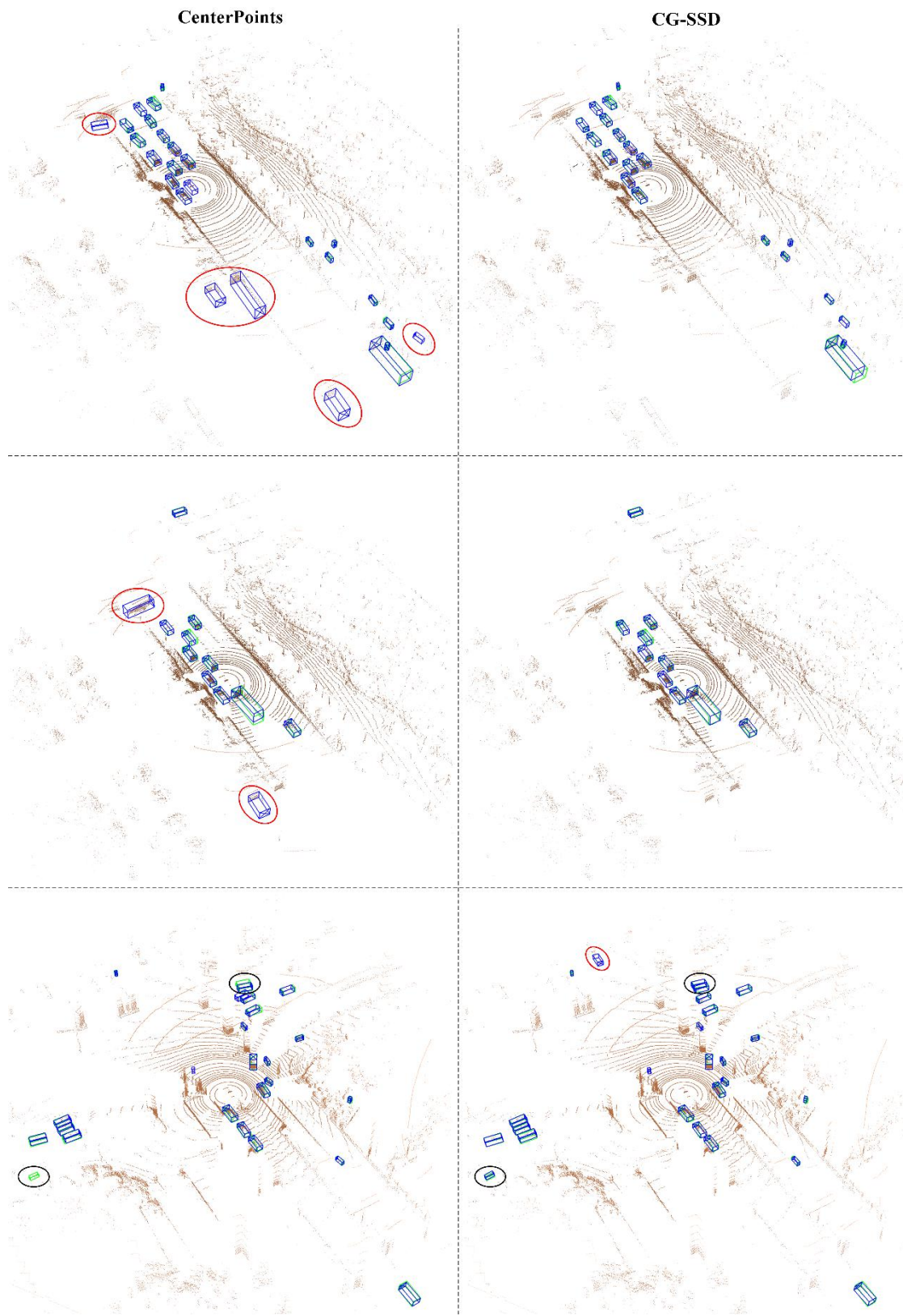

Figure 7. Visualization of object detection in ONCE. Left: CenterPoints(Yin et al., 2021), Right: CG-SSD. The ground truth and predicted bounding boxes are shown in green and blue respectively, the red ovals represent the falsely detected objects and the black ovals are the objects only detected by CG-SSD.

Table 3. Performance comparison on the ONCE validation set.

| Model | Vehicle (AP@50) | | | | Pedestrian (AP@50) | | | | Cyclist (AP@50) | | | | mAP |
|---|---|---|---|---|---|---|---|---|---|---|---|---|---|
| | overall | 0-30m | 30-50m | 50m-inf | overall | 0-30m | 30-50m | 50m-inf | overall | 0-30m | 30-50m | 50m-inf | |
| PointRCNN* (Shi et al., 2019) | 52.09 | 74.45 | 40.89 | 16.81 | 4.28 | 6.17 | 2.40 | 0.91 | 29.84 | 46.03 | 20.94 | 5.46 | 28.74 |
| PointPillars* (Lang et al., 2019) | 68.57 | 80.86 | 62.07 | 47.04 | 17.63 | 19.74 | 15.15 | 10.23 | 46.81 | 58.33 | 40.32 | 25.86 | 44.34 |
| SECOND* (Yan et al., 2018) | 71.19 | 84.04 | 63.02 | 47.25 | 26.44 | 29.33 | 24.05 | 18.05 | 58.04 | 69.96 | 52.43 | 34.61 | 51.89 |
| PV-RCNN* (Shi et al., 2020) | **77.77** | **89.39** | **72.55** | **58.64** | 23.50 | 25.61 | 22.84 | 17.27 | 59.37 | 71.66 | 52.58 | 36.17 | 53.55 |
| CenterPoints* (Yin et al., 2021) | 66.79 | 80.10 | 59.55 | 43.39 | 49.90 | 56.24 | 42.61 | 26.27 | 63.45 | 74.28 | 57.94 | 41.48 | 60.05 |
| CG-SSD(ours) | 67.60 | 80.22 | 61.23 | 44.77 | **51.50** | **58.72** | **43.36** | **27.76** | **65.79** | **76.27** | **60.84** | **43.35** | **61.63** |

Note: *: represents the result of the model is from ONCE Benchmark

## 4.4. Ablation Study

### 4.4.1. Type of Corners

The 3D object on BEV is represented using a rectangle with a rotation angle and has four corners. CG-SSD directly predicts the coordinates of the center point on the z-axis and the height of the object. So, the 3D bounding boxes can be recovered based on the rectangular box on the BEV and the predicted height. As shown in Figure 4, each object's four corners are divided into three types: visible corner ($VC$), partly visible corners ($PVCL, PVCW$), and invisible corner ($IVC$). We analyze the impact of different types of corner detection on ONCE validation set and set the CenterPoints(Yin et al., 2021) model as the baseline.

As shown in Table 4, when CGAM selects the $VC$ and $PVCW$ as the supervision signal, CG-SSD gets similar performance with baseline (+0.00%mAP, -0.04%mAP). But when the $PVCL$ and $IVC$ are selected, the model gets significant improvement with baseline (+0.71%mAP, +0.89%mAP). The $PVCL$ and $IVC$ are farther away from the $VC$ and more difficult to be observed by the LiDAR. This result shows that corners that are harder to be observed by the sensor can guide the model to learn a more accurate 3D bounding box.

Table 4. Effects of different corners for auxiliary model on ONCE validation set

| Corner | Vehicle (AP@50) | | | | Pedestrian (AP@50) | | | | Cyclist (AP@50) | | | | mAP |
|---|---|---|---|---|---|---|---|---|---|---|---|---|---|
| | overall | 0-30m | 30-50m | 50m-inf | overall | 0-30m | 30-50m | 50m-inf | overall | 0-30m | 30-50m | 50m-inf | |
| Baseline | 66.79 | 80.10 | 59.55 | 43.39 | 49.90 | 56.24 | 42.61 | 26.27 | 63.45 | 74.28 | 57.94 | 41.48 | 60.05 |
| $VC$ | 65.47 | 79.49 | 58.77 | 42.34 | 49.76 | 57.88 | 40.11 | 25.39 | 64.93 | 76.17 | 60.11 | 42.43 | 60.05 |
| $PVCW$ | 65.60 | 79.90 | 58.52 | 42.37 | 50.04 | 57.25 | 42.27 | 26.57 | 64.38 | 75.65 | 58.48 | 42.34 | 60.01 |
| $PVCL$ | 66.86 | 80.13 | 59.69 | 42.37 | 50.83 | 57.86 | 42.90 | 27.73 | 64.58 | 75.34 | 60.33 | 42.02 | 60.76 |
| $IVC$ | 67.34 | 79.92 | 59.71 | 44.48 | 50.75 | 57.85 | 42.76 | 27.89 | 64.72 | 75.79 | 59.84 | 42.82 | **60.94** |

### 4.4.2. Number of Corners

The ablation study of the type of corners shows that different corners contribute differently to the model detection accuracy. To get better results, we test our model with the different number of corners in the auxiliary module. CenterPoints(Yin et al., 2021) is also selected as the baseline. As shown in Table 5, when one corner ($IVC$) is selected as the supervision signal of the CGAM, CG-SSD shows better capacity than the baseline with +0.89%mAP. When two corners (*IVC* and *PVCL*) are added to CGAM, the results of the model are improved compared to when one corner is added(+0.18%mAP). When all corners (*IVC+PVCL+PVCW+VC*) are selected, the number of detection objects of the CGAM increases the difficulty of network convergence. Therefore, the model does not achieve higher detection results due to the increase in the number of detected corners. CG-SSD model gets the best result when three corners (*IVC*, *PVCL,* and *PVCW*) are selected as the detection object (+1.58%mAP vs baseline).

Table 5. Effects of different numbers of corners for auxiliary model on ONCE validation set

| Number of Corners | Vehicle (AP@50) | | | | Pedestrian (AP@50) | | | | Cyclist (AP@50) | | | | mAP |
|---|---|---|---|---|---|---|---|---|---|---|---|---|---|
| | overall | 0-30m | 30-50m | 50m-inf | overall | 0-30m | 30-50m | 50m-inf | overall | 0-30m | 30-50m | 50m-inf | |
| Baseline | 66.79 | 80.10 | 59.55 | 43.39 | 49.90 | 56.24 | 42.61 | 26.27 | 63.45 | 74.28 | 57.94 | 41.48 | 60.05 |
| One* | 67.34 | 79.92 | 59.71 | 44.48 | 50.75 | 57.85 | 42.76 | 27.89 | 64.72 | 75.79 | 59.84 | 42.82 | 60.94 |
| Two* | 67.17 | 80.02 | 60.33 | 43.99 | 50.57 | 57.77 | 42.63 | 27.20 | 65.63 | 76.25 | 60.68 | 42.77 | 61.12 |
| Three* | 67.60 | 80.22 | 61.23 | 44.77 | 51.50 | 58.72 | 43.36 | 27.76 | 65.79 | 76.27 | 60.84 | 43.35 | **61.63** |
| Four* | 68.53 | 79.97 | 61.18 | 46.89 | 49.38 | 56.64 | 41.46 | 26.20 | 65.39 | 76.45 | 60.35 | 44.09 | 61.10 |

*Note*: One: *IVC*, Two: *IVC + PVCL*, Three: *IVC + PVCL + PVCW*, Four: *IVC + PVCL + PVCW + VC*

### 4.4.3. Effects of Different Corner Auxiliary Signal

To select the most beneficial supervision signal in the CGAM, we select the corners classification and position information as the output of the auxiliary module to improve the detection accuracy of the model. As shown in Table 6, similar to the previous model(Hu et al., 2021; Wang et al., 2020) results, when only the corners classification information is added, the detection accuracy of the model will be improved by +0.34%mAP compared with CenterPoints(Yin et al., 2021). The experiment also shows that adding the corners offset is more beneficial to the detection accuracy of the model with +1.24%mAP. The ablation study proves that adding the classification and regression signals of the corners can guide the network to learn more accurate center point positions and bounding boxes of the target.

Table 6. Effects of different corner auxiliary signal on ONCE validation set

| Method | Vehicle (AP@50) | | | | Pedestrian (AP@50) | | | | Cyclist (AP@50) | | | | mAP |
|---|---|---|---|---|---|---|---|---|---|---|---|---|---|
| | overall | 0-30m | 30-50m | 50m-inf | overall | 0-30m | 30-50m | 50m-inf | overall | 0-30m | 30-50m | 50m-inf | |
| Baseline | 66.79 | 80.10 | 59.55 | 43.39 | 49.90 | 56.24 | 42.61 | 26.27 | 63.45 | 74.28 | 57.94 | 41.48 | 60.05 |
| w/ classification | 66.65 | 79.44 | 59.28 | 44.10 | 49.12 | 56.19 | 41.90 | 25.23 | 65.41 | 76.46 | 60.31 | 44.23 | 60.39 |
| w/ classification&offset | 67.60 | 80.22 | 61.23 | 44.77 | 51.50 | 58.72 | 43.36 | 27.76 | 65.79 | 76.27 | 60.84 | 43.35 | **61.63** |

### 4.4.4. Extension to BEV-based Models

The CGAM consists of a series of 2D convolutional modules and is easily extended to any 3D detector that uses a BEV map to detect objects. To verify it can play a plug-in to other models, we select PointPillars(Lang et al., 2019), SECOND(Yan et al., 2018), and PV-RCNN(Shi et al., 2020) to test on ONCE and Waymo validation set（based on 20% training data）. As shown in Table 7 and 8, on ONCE dataset, the auxiliary module can bring +3.37%~+7.05% mAP to the original detector. On Waymo dataset, enhanced models outperform the original model with +1.31%~+14.23% APH in L1 level and +1.17%~+12.93% APH in L2 level. Especially, for pedestrian and cyclist detection, our corner detection module has more advantages. These results show that after adding the CGAM, all three detectors get better performance on two datasets. Further, it can be extended as a plug-in to most 3D detectors that use a BEV map.

Table 7. Extension to some BEV map-based models on ONCE validation set

| Model | AP@50 | | | |
| --- | --- | --- | --- | --- |
| | Vehicle | Pedestrian | Cyclist | mAP |
| PointPillars* (Lang et al., 2019) | 68.57 | 17.63 | 46.81 | 44.34 |
| +auxiliary module | **70.45(+1.88)** | **18.97(+2.34)** | **53.71(+6.9)** | **47.71(+3.37)** |
| SECOND* (Yan et al., 2018) | **71.19** | 26.44 | 58.04 | 51.89 |
| +auxiliary module | 71.10(-0.09) | **38.27(+11.83)** | **61.89(+3.85)** | **57.09(+5.2)** |
| PV-RCNN* (Shi et al., 2020) | 77.77 | 23.50 | 59.37 | 53.55 |
| +auxiliary module | **79.20(+1.43)** | **37.19(+13.69)** | **65.39(+6.02)** | **60.60(+7.05)** |

Note: * represents the result is from ONCE dataset paper(Mao et al., 2021).

Table 8. Extension to some BEV map-based models on Waymo validation set

| Model | Vehicle/APH | | Pedestrian/APH | | Cyclist/APH | |
| --- | --- | --- | --- | --- | --- | --- |
| | L1 | L2 | L1 | L2 | L1 | L2 |
| PointPillars* (Lang et al., 2019) | 69.83 | 61.64 | 46.32 | 40.64 | 51.75 | 49.80 |
| +auxiliary module | **71.87 (+2.04)** | **63.93 (+2.29)** | **60.55 (+14.23)** | **53.57 (+12.93)** | **63.86 (+12.11)** | **61.51 (+11.71)** |
| SECOND* (Yan et al., 2018) | 70.34 | 62.02 | 54.24 | 47.49 | 55.62 | 53.53 |
| +auxiliary module | **71.65 (+1.31)** | **63.78 (+1.76)** | **58.01 (+3.77)** | **51.03 (+3.54)** | **63.31 (+7.69)** | **61.01 (+7.48)** |
| PV-RCNN* (Shi et al., 2020) | 74.74 | 66.80 | 61.24 | 53.95 | 64.25 | 61.82 |
| +auxiliary module | **76.53 (+1.79)** | **67.97 (+1.17)** | **66.62 (+5.38)** | **58.36 (+4.41)** | **69.63 (+5.38)** | **67.12 (+5.30)** |

Note: * represent the results are from the open-source project OpenPCDet (https://github.com/open-mmlab/OpenPCDet).

## 4.5. Discussion

There are a large number of LiDAR points around the visible corner of objects, and it is easily detected by the network. Therefore, adding the learning of it to the auxiliary module cannot improve the detection accuracy of the network. When we add the partly visible corners to the auxiliary module, our model gets better results with the selected corner is further away from the visible corner (section 4.4.1). The ablation study of the types of corners shows that CGAM can improve detection accuracy, and learning invisible corner information is more important for the detector.

The object is represented by a rectangle in the BEV map, and the three corners can determine the shape of a rectangle. Therefore, the model can perceive the complete target information and achieve the best detection results when three corners are selected as the supervision signals (section 4.4.2). Selecting four corners as the detection object cannot achieve better accuracy, which will not only generate redundant information but also increase the complexity and computational burden of the model. Furthermore, selecting the corner effective supervision signal is also a critical problem in the CGAM. The corner classification and offset value are the key features and they can bring a significant improvement in detection accuracy(section 4.4.3). The precise location information of the corners can play an important guiding role when the network predicts the bounding boxes.

The CGAM consists of a series of CBR blocks, and it is easy to be extended to other models which use the BEV feature map to predict objects. The CGAM is added to three anchor-based models. Due to the CGAM can provide more information of the object corners and guide the network to predict more accurate 3D bounding boxes, the improved models get better performance than original models on ONCE and Waymo datasets (section 4.4.4). Especially, because the CGAM provide different corners' location which can implicitly express the position of the center point, the improved anchor-based models show better capacity on the small objects (pedestrian and cyclist).

The anchor-free model has a good ability to detect and localize small targets. However, with the BEV feature map representation, due to the size limitation of the grid, small targets occupy only one or two pixels in the BEV map. Therefore, it will bring the problem of small target shape information loss. Our proposed corner detection auxiliary module can provide accurate corner information of the target, which greatly reduces the loss of target information and thus improves the target detection accuracy. For pedestrian and cyclist detection, our model shows more significant improvement compared with other models.

## 5. Conclusion

Due to the placement position of the LiDAR sensor, the scanning angle, and the occlusion between the targets, the LiDAR sensor cannot obtain complete target point clouds data. It is very difficult to predict and regress accurate 3D bounding boxes of objects from incomplete object point clouds. In this paper, we have proposed a one-stage corner-guided anchor-free 3D object detection model (CG-SSD). The CG-SSD model innovatively takes the corner information of the object into the training and inference of the network, to guide the network to obtain more accurate 3D bounding boxes from the limited point clouds. An adaptive corner classification method is designed according to the limited number of object points. Different from the previous model to obtain the 3D bounding box of the target by detecting the center point, we innovatively added the corner prediction module CGAM in CG-SSD. The experiment on the ONCE dataset shows that our network achieves better performance than other state-of-the-art models (CenterPoints(Yin et al., 2021) and PV-RCNN(Shi et al., 2020)) for supervised 3D object detection using single frame point clouds data. In addition, the results on ONCE and Waymo data show that our proposed CGAM can be extended to other 3D detectors which use BEV map as a plug-in for boosting the performances. The results of this paper demonstrate the importance of corner detection in 3D object detection. In our future work, we will focus on building a corner point-based model instead of center-based ones to fuse different sensor modalities to detect and track 3D objects continuously in space and time for fully self-driving cars.